\documentclass[conference]{IEEEtran}
\IEEEoverridecommandlockouts
\usepackage[compress]{cite}
\usepackage{balance}
\usepackage{amsmath,amssymb,amsfonts}
\usepackage{algorithmic}
\usepackage{algorithm}
\usepackage{graphicx}
\usepackage{subcaption}
\usepackage{textcomp}
\usepackage{tikz}
\usepackage[utf8]{inputenc} % upućujete LaTeX da je izvorni kod zapisan u utf8 unicode encodingu
\usepackage[T1]{fontenc} % upućujete LaTeX da koristi font family T1; ovaj redak je nužan zbog slova đ i Đ
\def\BibTeX{{\rm B\kern-.05em{\sc i\kern-.025em b}\kern-.08em
    T\kern-.1667em\lower.7ex\hbox{E}\kern-.125emX}}

\begin{document}

\title{Stereo Event Lifetime and Disparity Estimation for Dynamic Vision Sensors\\
}

\author{Antea Hadviger, Ivan Marković, Ivan Petrović$^\ast$
\thanks{
This work has been supported by the FLAG-ERA JTC 2016 and the Ministry of Science and Education of the Republic of Croatia under the project \textit{Rethinking Robotics for the Robot Companion of the Future} (RoboCom++) and the Croatian Science Foundation under contract No. I-3137-2019. The research has been carried out within the activities of the \textit{Centre of Research Excellence for Data Science and Cooperative Systems} supported by the Ministry of Science and Education of the Republic of Croatia.
}
\thanks{
$^{\ast}$Authors are with the University of Zagreb
Faculty of Electrical Engineering and Computing, Laboratory for Autonomous Systems and Mobile Robotics, Croatia. {\{antea.hadviger, ivan.markovic, ivan.petrovic\}@fer.hr }
\vspace{2mm}

\noindent 978-1-7281-3605-9/19/\$31.00~\copyright2019 IEEE
}
}

\maketitle

\begin{abstract}
Event-based cameras are biologically inspired sensors that output asynchronous pixel-wise brightness changes in the scene called events.
They have a high dynamic range and temporal resolution of a microsecond, opposed to standard cameras that output frames at fixed frame rates and suffer from motion blur.
Forming stereo pairs of such cameras can open novel application possibilities, since for each event depth can be readily estimated; however, to fully exploit asynchronous nature of the sensor and avoid fixed time interval event accumulation, stereo event lifetime estimation should be employed.
In this paper, we propose a novel method for event lifetime estimation of stereo event-cameras, allowing generation of sharp gradient images of events that serve as input to disparity estimation methods.
Since a single brightness change triggers events in both event-camera sensors, we propose a method for single shot event lifetime and disparity estimation, with association via stereo matching.
The proposed method is approximately twice as fast and more accurate than if lifetimes were estimated separately for each sensor and then stereo matched.
Results are validated on real-world data through multiple stereo event-camera experiments.
\end{abstract}

\begin{IEEEkeywords}
event-based cameras, stereo vision, event lifetime estimation, disparity estimation
% \vspace{-0.3cm}
\end{IEEEkeywords}

%%%%%%%%%%%%%%%%%%%%%%%%%%%%%%%%%%%%%%%%%%%%%%%%%%%%%%%%%%%%%%%

\vspace{1mm}
\section{Introduction} \label{sec:intro}

\begin{figure}[t!]
%\begin{subfigure}{.25\textwidth}
%  \centering
%  % nedjelja_2019-04-28-11-49-50.bag
%  \frame{\includegraphics[width=.95\textwidth]{figures/komp00001249_2.png}}
%  \caption{}
%  %\label{fig:sfig1}
%\end{subfigure}%
\vspace{0.15cm}
\begin{subfigure}{.24\textwidth}
  \centering
  \frame{\includegraphics[width=.98\textwidth]{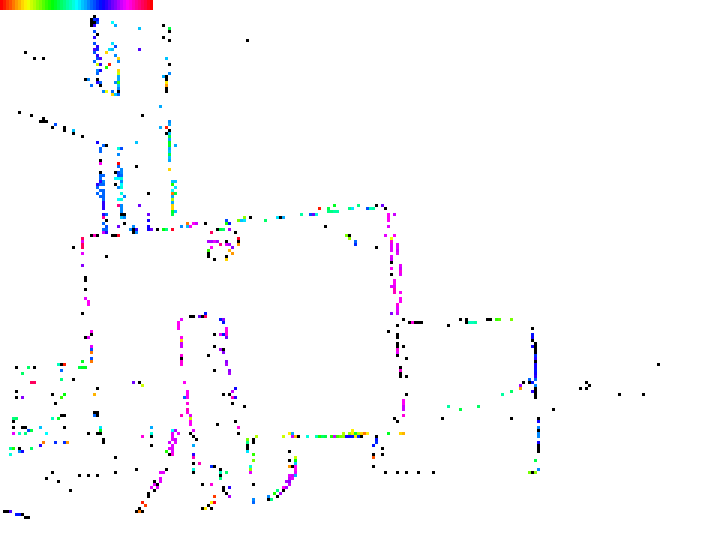}}
  \caption{Fixed accumulation interval}
  %\label{fig:sfig2}
\end{subfigure}
\vspace{0.15cm}
\begin{subfigure}{.24\textwidth}
  \centering
  \frame{\includegraphics[width=.98\textwidth]{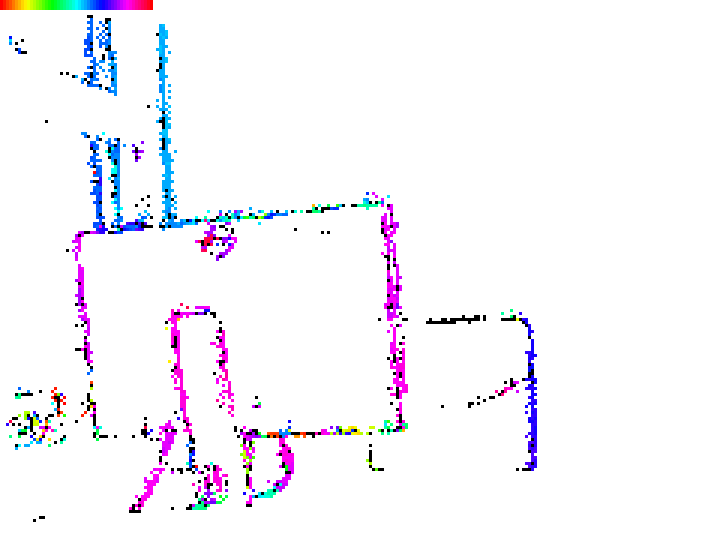}}
  \caption{Proposed method}
  %\label{fig:sfig2}
\end{subfigure}
\begin{subfigure}{.24\textwidth}
 \centering
 \frame{\includegraphics[width=.98\textwidth]{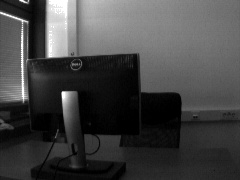}}
 \caption{Grayscale image}
 \label{fig:sfig3}
\end{subfigure}
\begin{subfigure}{.24\textwidth}
 \centering
 \vspace{0.2cm}
 \includegraphics[width=.98\textwidth]{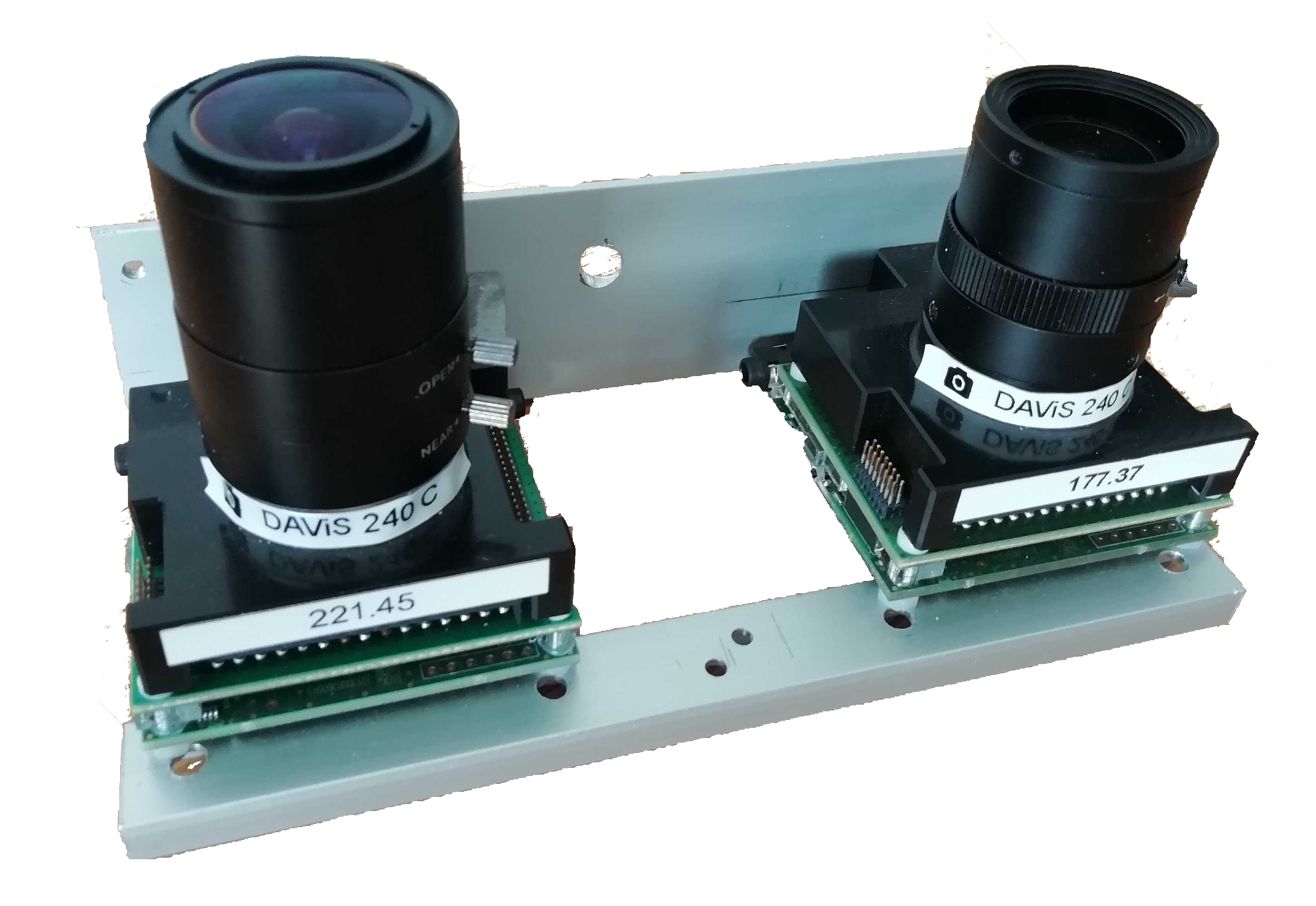}
 %\vspace{0.5cm}
 \caption{Experimental stereo setup}
 \label{fig:sfig4}
\end{subfigure}
\caption{Stereo DVS events -- using a fixed event accumulation time of 10 ms yields a sparse event disparity map prone to noise, while disparity obtained with the proposed method retains scene structure and  successfully filters noise. Color bar in the top left corner indicates disparity, i.e., depth. Black pixels indicate no available disparity.}
%\im{Predlažem da ovdje  stavimo ipak 4 slike, da 3. bude RGB slika.}}
\label{fig:exp2}
\vspace{-5mm}
\end{figure}

Event-based vision sensors \cite{delbruck2010activity}, such as the dynamic vision sensor (DVS) \cite{lichtsteiner2008128} and dynamic and active-pixel vision sensor (DAVIS) \cite{brandli2014240}, are relatively novel biologically inspired cameras that output pixel-wise changes in brightness intensity in the scene,  which are referred to as \textit{events} due to their asynchronous nature.
Since event-based vision sensors do not record the brightness intensity of an entire scene at a fixed frame rate, but offer a sparse data stream reporting only changes in the intensity, they have significantly lower power consumption and bandwidth requirements than standard cameras still widely used in computer vision.
Furthermore, event-based cameras operate with low latency and with a temporal resolution of a microsecond, thus avoiding the problem of motion blur that makes standard cameras unusable in highly dynamic scenarios.
Also, event-based cameras have the advantage of a high dynamic range (HDR) of 140 dB, compared to the 60 dB of standard cameras, making them an interesting sensor for tasks with challenging scene illumination and environmental conditions, often encountered by autonomous mobile robots.
A recent survey on event-based vision applications and methods can be found in \cite{gallego2019event} and an example of a processed DVS output can be seen in Fig.~\ref{fig:exp2}.

%sve_7669: 2976 1623, 18784550949 3256535

Standard computer vision algorithms that operate on image intensity frames cannot be directly applied to the fundamentally different event stream of event-based vision sensors.
%To fully exploit the potential of the sensor, it is required either to develop new specialized algorithms or adapt event data to accomodate the existing methods.
Recently, significant efforts have been made in the area of intensity image reconstruction from events \cite{rebecq2019events,wang2018event}; however, for some applications complex image intensity reconstruction is not needed.
If the incoming events are simply accumulated over time, they can be represented as a frame and used as an input to slightly altered standard image processing algorithms.
Accumulation can be done directly by choosing a fixed time interval and this approach has been used for tracking and optical flow estimation \cite{ni2012asynchronous,bardow2016simultaneous}.
Events can also be accumulated by choosing a certain number of events to be shown in each generated frame, as shown in \cite{vidal2018ultimate} for simultaneous localization and mapping (SLAM).
Since the frequency of incoming events depends on the amount of motion of the scene and the camera itself, an accumulation interval that is too long will produce blurred images, while a short interval will cause loss of structure in the scene, thus not allowing correct interpretation.
Given that, regardless of the strategy, event accumulation introduces latency and does not exploit fully the asynchronous nature and high temporal resolution of the sensor.
To overcome the aforementioned problems, the nature of the sensor implies that each event should be dynamically augmented with its duration and velocity.
Authors in \cite{mueggler2015lifetime} introduced the concept of an \textit{event lifetime} and propose to estimate it by leveraging optical flow \cite{benosman2014event}, thus allowing to identify a set of active events at any given time instance.
An event is considered active as long as the change in brightness that triggered it is visible by its respective pixel.
Using event lifetimes to identify the set of active events implicitly generates sharp gradient images at any point in time, thus effectively functioning as an edge detector.

Event-based vision sensors have also been proven successful in depth estimation via stereo matching.
Some event-driven methods rely on standard stereo matching algorithms by generating artificial event frames.
In \cite{schraml2010dynamic,kogler2011address} authors used a fixed event accumulation interval to generate artificial frames, while \cite{kogler2011event} generated different frames based on event timestamps.
A similar approach is to match events by comparing their local context by forming descriptors \cite{zou2016context}.
%Another approach are "frameless" methods that use temporal correlation of events across sensors \cite{kogler2011event}.
However, opposed to choosing a fixed time event accumulation interval, enhancing the event stream with event lifetimes yields more accurate disparity maps \cite{zou2017disparity}.
Therein the authors proposed to run stereo lifetime estimation separately for each sensor and subsequently estimate the disparity, which requires either using multiple threads or introduces latency caused by lifetime computation.
Since the two sensors are mostly observing the same scene, they will also emit events triggered by the same brightness changes, thus offering the possibility to exploit the fact that the corresponding events should also have the same lifetimes.

In this paper we propose a novel method for event lifetime and disparity estimation in stereo event-cameras.
Since events in a stereo setup are caused by the same change in brightness, we couple stereo event disparity and lifetime estimation in order to calculate them efficiently in a single shot.
The resulting disparity estimation method is computationally more efficient and accurate in comparison to the method that decouples event lifetime and disparity estimation \cite{zou2017disparity}.
The proposed approach is validated on real-world data through multiple stereo event-camera experiments.
The algorithm runs in a single thread, making it appropriate for energy-efficient applications in autonomous systems and mobile robotics.

The rest of the paper is organized as follows.
In Section~\ref{sec:lifetime} we describe the event lifetime estimation method proposed in \cite{mueggler2015lifetime} based on optical flow and describe the local plane fitting algorithm.
Our novel method for stereo event lifetime estimation together with disparity calculation is presented in Section~\ref{sec:method}.
Finally, we present the experimental results in Section~\ref{sec:exp} and conclude with final remarks in Section~\ref{sec:conclusion}.

%%%%%%%%%%%%%%%%%%%%%%%%%%%%%%%%%%%%%%%%%%%%%%%%%%%%%%%%%%%%%%%%%%%%%%

\vspace{0.3cm}
\section{Optical Flow Based Event Lifetime Estimation} \label{sec:lifetime}

Event lifetime estimation was proposed in \cite{mueggler2015lifetime} as a way to identify the set of active events at any given time instance, resulting in frames of accumulated events with clear structure and filtered noise that can be used as an input to standard computer vision algorithms for further processing.
Herein we provide its description for completeness as it is an important part of the proposed stereo event lifetime and disparity estimation method.

\subsection{Method Overview}

\begin{figure}[!t]
	\centering
  	\vspace{-0.5cm}
	\includegraphics[width=0.4\textwidth]{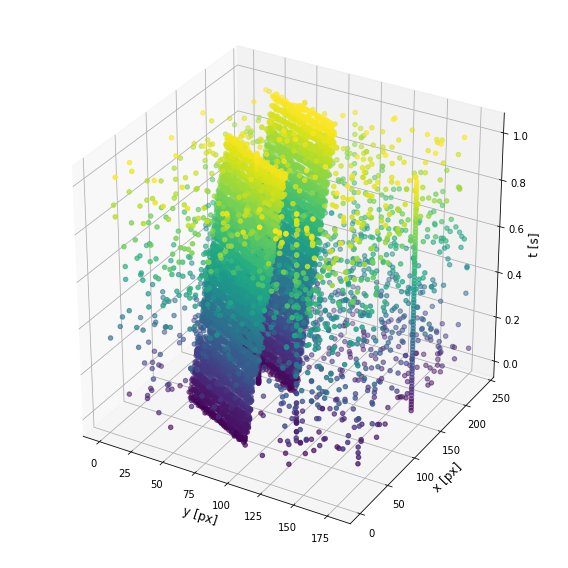}
	\caption{Surface of active events for a moving line depicting events from both the left and right camera, where outliers are caused by sensor noise. Color gradient represents the timestamp for easier visibility.}
	\label{fig:sae}
\end{figure}

The surface of active events (SAE) is a three-dimensional structure defined in the spatio-temporal domain where coordinates of each pixel of the sensor are mapped to the timestamp of the last event that occurred on that position (Figure \ref{fig:sae}).
Formally, SAE can be described using a function $S : \mathbb{R}^{2} \rightarrow \mathbb{R}, S(\mathbf{p})=t,$
%$\Sigma_{e} : \mathbb{R}^{2} \rightarrow \mathbb{R}, t=\Sigma_{e}(\mathbf{p})$,
where $\mathbf{p} = (x, y)^{\top}$ is the pixel position and $t$ is the corresponding timestamp.
% Therefore, each SAE point is represented by its location $\mathbf{p}$ and timestamp $S(\mathbf{p})$. %$S(\mathbf{p})=\left(x, y, \Sigma_{e}(x, y)\right)^{\top}$.
%
The first order Taylor expansion gives the planar approximation of the SAE at a pixel location $\mathbf{p}$:
\begin{equation}
S(\mathbf{p}+\Delta \mathbf{p}) \approx S(\mathbf{p})+\left(S_{x}(\mathbf{p}), S_{y}(\mathbf{p})\right) \Delta \mathbf{p},
\end{equation}
where $S_{x}$ and $S_{y}$ denote the first partial derivatives, $\frac{\partial S}{\partial x}$ and $\frac{\partial S}{\partial y}$,
%$\frac{\partial S}{\partial x}=\left(1,0, \frac{\partial \Sigma_{c}}{\partial x}\right)^{\top}$ and $\frac{\partial S}{\partial y}=\left(0,1, \frac{\partial \Sigma_{e}}{\partial y}\right)^{\top}$,
respectively.
As proposed in \cite{mueggler2015lifetime}, lifetime $\tau$ of the event at the point $(\mathbf{p}, t)$ is defined as the first order approximation of the maximum temporal increment of $S$ for a displacement of one pixel.
Therefore, the lifetime $\tau$ corresponds to the maximum amount of time before the brightness change at the current pixel will trigger a new event in a neighboring pixel.
In other words, only one event caused by a single change in brightness should be considered active at some point in time; this can formally be expressed as:
\begin{equation}
\tau(\mathbf{p})=\max \{\Delta t\}, \text { subject to } \|\Delta \mathbf{p}\|=1,
\end{equation}
where $\Delta t = S(\mathbf{p}+\Delta \mathbf{p}) - S(\mathbf{p})$. Substitution further yields $\Delta t = \left(S_{x}(\mathbf{p}), S_{y}(\mathbf{p})\right) \Delta \mathbf{p} = \nabla S(\mathbf{p}) \Delta \mathbf{p}$.
Finally, lifetime $\tau$ can be expressed as:
\begin{equation}
\tau(\mathbf{p})=\max \{\nabla S(\mathbf{p}) \Delta \mathbf{p}\}, \text { subject to } \|\Delta \mathbf{p}\|=1.
\end{equation}
Time $t$ being a naturally increasing function, $S(\mathbf{p})$ is a monotonically increasing function of $\mathbf{p}$, thus having nonzero gradient at any point.
This allows us to use the inverse function theorem around the location $\mathbf{p}$, as shown in \cite{benosman2014event}:
\begin{equation}
\begin{aligned}
\frac{\partial S}{\partial x}\left(x, y_{0}\right)=\frac{1}{v_{x}\left(x, y_{0}\right)}, \\
\frac{\partial S}{\partial y}\left(x_{0}, y\right)=\frac{1}{v_{y}\left(x_{0}, y\right)}.
\end{aligned}
\end{equation}
The gradient can thus be expressed as:
\begin{equation} \label{eq:1}
\nabla S(\mathbf{p}) = \left(\frac{1}{v_{x}(\mathbf{p})}, \frac{1}{v_{y}(\mathbf{p})}\right)^{\top},
\end{equation}
showing its relation to the optical flow velocities, $v_{x}$ and $v_{y}$.
Assumption of constant velocities is equivalent to the local planar approximation of the SAE. %\im{Na što se ova rečenica veže?}
The maximum temporal increment $\Delta t$ occurs if the displacement $\Delta \mathbf{p}$ matches the unit vector in the direction of gradient $\Delta \mathbf{p} = \nabla S(\mathbf{p})/\|\nabla S(\mathbf{p})\|$.
Therefore, %\im{check with ah},
\begin{equation}
\tau(\mathbf{p}) = \nabla S(\mathbf{p})\frac{\nabla S(\mathbf{p})}{\|\nabla S(\mathbf{p})\|} = \sqrt{v_{x}^{-2} + v_{y}^{-2}}.
\end{equation}
The normal of the tangent plane to the SAE at the point $\mathbf{p}$ is given by
\begin{equation} \label{eq:2}
\begin{split}
\mathbf{n}(\mathbf{p}) &\propto (1, 0, S_{x}(\mathbf{p}))^{\top} \times (0, 1, S_{y}(\mathbf{p}))^{\top}\\
 &= (-S_{x}(\mathbf{p}), -S_{y}(\mathbf{p}), 1)^{\top}.
\end{split}
\end{equation}
Furthermore, combination of \eqref{eq:1} and \eqref{eq:2} yields $\mathbf{n}(\mathbf{p})~\propto~(-v_{x}^{-1}, -v_{y}^{-1}, 1)^{\top}$, and naming the normal coordinates as $\mathbf{n}(\mathbf{p})=(n_1, n_2, n_3)^{\top}$ implies $-v_{x}^{-1} = \frac{n_1}{n_3}$ and $-v_{y}^{-1} = \frac{n_2}{n_3}$.
Components of the SAE tangent plane normal $(n_1, n_2, n_3)^{\top}$ are extracted by fitting a plane to the data using an optimization technique (see Section \ref{sec:plane}) that finally enables computation of the lifetime $\tau(\mathbf{p})$ as:

\begin{equation}
\tau(\mathbf{p}) = \frac{1}{n_3} \sqrt{n_1^2 + n_2^2}.
\end{equation}

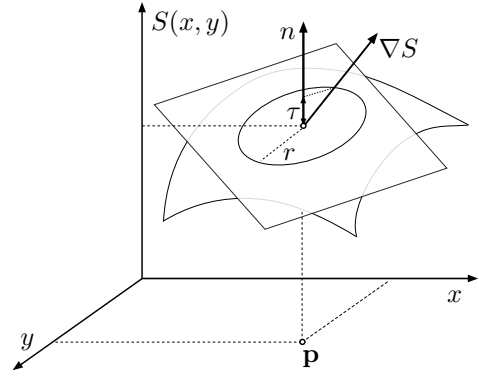
\begin{figure}[!t]
	\centering
	\definecolor{cffffff}{RGB}{255,255,255}

\begin{tikzpicture}[y=0.80pt, x=0.80pt, yscale=-1.000000, xscale=1.000000, inner sep=0pt, outer sep=0pt]
  \path[fill=black,nonzero rule] (553.0927,804.2692) -- (548.2392,802.4088) --
    (548.2148,806.0612) -- cycle(553.0927,804.2692);
  \path[draw=black,line join=round,line cap=round,miter limit=10.00,line
    width=0.560pt] (336.8905,844.9920) -- (394.2991,804.2425);
  \path[fill=black,nonzero rule] (333.0041,847.6427) -- (338.0334,846.3097) --
    (335.9143,843.3312) -- cycle(333.0041,847.6427);
  \path[draw=black,line join=round,line cap=round,miter limit=10.00,line
    width=0.560pt] (394.2158,804.3256) -- (394.2158,678.2953);
  \path[fill=black,nonzero rule] (394.2158,673.4271) -- (392.3896,678.2953) --
    (396.0371,678.2953) -- cycle(394.2158,673.4271);
  \begin{scope}[cm={{1.25,0.0,0.0,1.25,(-109.27544,167.96327)}}]
    \begin{scope}[fill=black]
    \end{scope}
  \end{scope}
  \begin{scope}[cm={{1.25,0.0,0.0,1.25,(-109.27544,167.96327)}}]
    \begin{scope}[fill=black]
    \end{scope}
  \end{scope}
  \begin{scope}[cm={{1.25,0.0,0.0,1.25,(-109.27544,167.96327)}}]
    \begin{scope}[fill=black]
    \end{scope}
  \end{scope}
  \begin{scope}[cm={{1.25,0.0,0.0,1.25,(-109.27544,167.96327)}}]
    \begin{scope}[fill=black]
    \end{scope}
  \end{scope}
  \begin{scope}[cm={{1.25,0.0,0.0,1.25,(-109.27544,167.96327)}}]
    \begin{scope}[fill=black]
    \end{scope}
  \end{scope}
  \begin{scope}[cm={{1.25,0.0,0.0,1.25,(-109.27544,167.96327)}}]
    \begin{scope}[fill=black]
    \end{scope}
  \end{scope}
  \begin{scope}[cm={{1.25,0.0,0.0,1.25,(-109.27544,167.96327)}},fill=black]
  \end{scope}
  \begin{scope}[cm={{1.25,0.0,0.0,1.25,(-109.27544,167.96327)}}]
    \begin{scope}[fill=black]
    \end{scope}
  \end{scope}
  \begin{scope}[cm={{1.25,0.0,0.0,1.25,(-109.27544,167.96327)}}]
    \begin{scope}[fill=black]
    \end{scope}
  \end{scope}
  \begin{scope}[cm={{1.25,0.0,0.0,1.25,(-109.27544,167.96327)}},fill=black]
  \end{scope}
  \begin{scope}[cm={{1.25,0.0,0.0,1.25,(-109.27544,167.96327)}}]
    \begin{scope}[fill=black]
    \end{scope}
  \end{scope}
  \begin{scope}[cm={{1.25,0.0,0.0,1.25,(-109.27544,167.96327)}}]
    \begin{scope}[fill=black]
    \end{scope}
  \end{scope}
  \begin{scope}[cm={{1.25,0.0,0.0,1.25,(-109.27544,167.96327)}},fill=black]
  \end{scope}
  \begin{scope}[cm={{1.25,0.0,0.0,1.25,(-109.27544,167.96327)}}]
    \begin{scope}[fill=black]
    \end{scope}
  \end{scope}
  \begin{scope}[cm={{1.25,0.0,0.0,1.25,(-109.27544,167.96327)}},fill=black]
  \end{scope}
  \begin{scope}[cm={{1.25,0.0,0.0,1.25,(-109.27544,167.96327)}},fill=black]
  \end{scope}
  \begin{scope}[cm={{1.25,0.0,0.0,1.25,(-109.27544,167.96327)}}]
    \begin{scope}[fill=black]
    \end{scope}
  \end{scope}
  \begin{scope}[cm={{1.25,0.0,0.0,1.25,(-109.27544,167.96327)}}]
    \begin{scope}[fill=black]
    \end{scope}
  \end{scope}
  \begin{scope}[cm={{1.25,0.0,0.0,1.25,(-109.27544,167.96327)}}]
    \begin{scope}[fill=black]
    \end{scope}
  \end{scope}
  \path[draw=black,fill=black,line join=miter,line cap=butt,miter limit=4.00,fill
    opacity=0.000,even odd rule,line width=0.357pt] (548.5206,731.5454) ..
    controls (527.6232,720.8613) and (507.6703,708.8710) .. (488.9926,706.1405) ..
    controls (472.0130,703.6581) and (456.5528,703.7252) .. (442.4491,712.7484) ..
    controls (421.6861,726.0319) and (406.1280,746.2643) .. (404.3506,777.0744) ..
    controls (415.0207,770.4593) and (427.5939,766.6363) .. (440.7453,766.3325) ..
    controls (445.4279,766.2244) and (450.1837,766.5623) .. (454.9531,767.3792) ..
    controls (469.4362,769.8601) and (484.0926,777.3868) .. (495.7341,784.5632);
  \path[draw=black,fill=cffffff,line join=miter,line cap=butt,miter
    limit=4.00,even odd rule,line width=0.358pt] (495.7428,784.6477) .. controls
    (492.5777,773.8210) and (500.1369,758.9438) .. (508.1401,748.8994) .. controls
    (511.2433,745.0046) and (514.4132,741.8365) .. (517.0506,739.9128) .. controls
    (529.6505,730.7225) and (536.9415,732.5127) .. (548.5194,731.6695);
  \path[cm={{0.66376,0.74794,-0.9478,0.31886,(0.0,0.0)}},draw=black,fill=cffffff,opacity=0.850,miter
    limit=4.00,line width=0.339pt,rounded corners=0.0000cm] (881.9557,104.8838)
    rectangle (960.7672,195.5770);
  \path[draw=black,dash pattern=on 0.96pt off 0.96pt,line join=miter,line
    cap=butt,miter limit=5.33,even odd rule,line width=0.320pt]
    (469.8802,772.7625) -- (469.8802,802.4279) -- (469.8802,834.8541);
  \path[draw=black,dash pattern=on 0.96pt off 0.96pt,line join=miter,line
    cap=butt,miter limit=5.33,even odd rule,line width=0.320pt]
    (469.6298,834.3533) -- (351.8937,834.3533);
  \path[draw=black,dash pattern=on 0.96pt off 0.96pt,line join=miter,line
    cap=butt,miter limit=6.00,even odd rule,line width=0.320pt]
    (469.9062,732.0219) -- (394.1745,732.0219);
  \path[draw=black,dash pattern=on 0.96pt off 0.96pt,line join=round,line
    cap=round,miter limit=5.33,line width=0.320pt] (469.9985,834.2947) --
    (510.8034,805.4576);
  \path[fill=black,line join=miter,line cap=butt,line width=0.800pt]
    (336.6864,838.0303) node[above right] (text3650) {$y$};
  \path[fill=black,line join=miter,line cap=butt,line width=0.800pt]
    (538.4791,815.3908) node[above right] (text3650-3) {$x$};
  \path[fill=black,line join=miter,line cap=butt,line width=0.800pt]
    (470.0392,846.0248) node[above right] (text3650-7-3-7-3) {$\mathbf{p}$};
  \path[fill=black,line join=miter,line cap=butt,line width=0.800pt]
    (398.4638,689.6477) node[above right] (text3650-7-3-9) {$S(x,y)$};
  \path[draw=black,line join=round,line cap=round,miter limit=10.00,line
    width=0.618pt] (394.7936,804.2439) -- (548.4631,804.2439);
  \path[draw=black,fill=cffffff,miter limit=4.00,line width=0.561pt]
    (470.0350,834.2519) ellipse (0.0335cm and 0.0328cm);
  \path[cm={{-0.91261,0.40883,-0.63554,-0.77207,(0.0,0.0)}},draw=black,line
    join=miter,line cap=butt,miter limit=4.00,even odd rule,line width=0.384pt]
    (106.2903,-891.9838) ellipse (0.8653cm and 0.4899cm);
  \path[fill=black,nonzero rule] (505.7052,687.9899) -- (500.6930,690.9693) --
    (503.9336,693.4606) -- cycle(505.7052,687.9899);
  \path[draw=black,line join=round,line cap=round,miter limit=10.00,line
    width=0.718pt] (470.5885,732.2430) -- (503.0793,691.4058);
  \path[fill=black,nonzero rule] (470.5242,682.5340) -- (468.4255,687.8941) --
    (472.5458,687.9211) -- cycle(470.5242,682.5340);
  \path[draw=black,line join=miter,line cap=butt,miter limit=4.00,even odd
    rule,line width=0.890pt] (470.4817,731.8207) -- (470.4817,686.4984);
  \path[draw=black,dash pattern=on 0.51pt off 0.51pt,line join=miter,line
    cap=butt,miter limit=4.00,even odd rule,line width=0.170pt]
    (484.7982,714.0547) -- (470.5397,718.2948);
  \path[xscale=0.904,yscale=1.106,fill=black,line join=miter,line cap=butt,line
    width=0.800pt] (511.6209,658.7886) node[above right] (text3650-7-3-7) {$\tau$};
  \path[draw=black,dash pattern=on 0.81pt off 0.81pt,line join=miter,line
    cap=butt,miter limit=4.00,even odd rule,line width=0.271pt]
    (469.7499,733.4289) .. controls (450.4556,748.4735) and (450.4556,748.4735) ..
    (450.4556,748.4735);
  \path[cm={{1.00469,0.0251,-0.00081,0.99532,(0.0,0.0)}},fill=black,line
    join=miter,line cap=butt,line width=0.800pt] (459.3216,740.2825) node[above
    right] (text3650-7-3-7-3-6) {$r$};
  \path[fill=black,line join=miter,line cap=butt,line width=0.800pt]
    (506.1664,699.3982) node[above right] (text3650-7-3) {$\nabla S$};
  \path[fill=black,line join=miter,line cap=butt,line width=0.800pt]
    (459.3774,690.9847) node[above right] (text3650-7-3-1) {$n$};
  \path[fill=black,nonzero rule] (470.4731,717.9030) -- (469.0497,721.3283) --
    (471.8442,721.3455) -- cycle(470.4731,717.9030);
  \path[fill=black,nonzero rule] (470.5092,731.9933) -- (471.9326,728.5680) --
    (469.1381,728.5508) -- cycle(470.5092,731.9933);
  \path[draw=black,fill=cffffff,miter limit=4.00,line width=0.561pt]
    (470.4912,731.9987) ellipse (0.0335cm and 0.0328cm);

\end{tikzpicture}
	\caption{Principle of optical flow computation using SAE. Planar approximation of SAE is used to compute lifetimes of the events. Radius $r$ equals $1$.}
	\label{fig:flow}
\end{figure}

\subsection{Local plane-fitting Algorithm} \label{sec:plane}

A plane-fitting algorithm is needed to calculate the parameters of the SAE tangent plane to further estimate the lifetime of the given event at the location $\mathbf{p}$.
In \cite{benosman2014event}, it was proposed to use an $N \times N \times 2 \Delta t$ window centered around the current event to estimate the local plane, including the events from the future.
To avoid the introduced $\Delta t$ latency, we use only the past events in the $N \times N$ spatial window based on the local smoothness assumption, as suggested by \cite{mueggler2015lifetime}.
Even though therein they completely eliminate the $\Delta t$ parameter and the need to tune it, we use $\Delta t = 100\, \mathrm{ms}$ as a time frame in which the past events in the spatial window are considered to ensure that the events that are too old and  unrelated to recent events do not affect the estimated tangent plane negatively. %\im{raspraviti}.

Fitting the plane using the RANSAC algorithm suppresses noise events introduced by the sensor and ensures robustness.
First, a plane is fitted to a set of hypothetical inliers, which include the current event and two random past events in the spatio-temporal window.
All other past events are then tested against the fitted plane. They are considered as inliers if their distance to the plane is below the threshold $\mu$.
If less than $m$ inliers are found, the process is repeated from the beginning with a new hypothetical set of inliers.
The event is declared as noise if the condition for the minimum number of inliers is not met in the predefined maximum number of iterations.
Otherwise, the plane is reestimated by fitting to all the inliers using the least squares minimization:
\begin{equation}
\mathbf{n}_{\mathrm{L S}}=\underset{\mathbf{n}}{\arg \min }\|\mathbf{A} \mathbf{n}-\mathbf{b}\|^{2},
\end{equation}
where $\mathbf{A}$ is the $n \times 3$ matrix of the $n$ SAE points $(x_i, y_i, t_i)$ identified as inliers, and $\mathbf{b}=(1 \cdots 1)^{\top}$.

Assumption of locally constant velocity allows for prediction of the future events timestamp, as suggested in \cite{mueggler2015lifetime}, wherein authors proposed to use the predicted timestamp to compute the regularization factor depending on the absolute difference of the predicted timestamp $\hat{t_i}$ and the actual timestamp $t_i$ of the incoming events:

\begin{equation}
\Delta t_{\mathrm{err}} = |t_i - \hat{t_i}|.
\end{equation}
That method performs plane fitting based on the RANSAC algorithm for each event, even if the predicted plane normal is considered very reliable.
The more the predicted and actual timestamp are close, the more the plane normal obtained from the previous events is considered reliable.
Our method, however, avoids least squares optimization for the events whose timestamps closely match the predicted ones.
In that case, we assume that the normal of the current event tangent plane is equal to the predicted normal, saving computational time while retaining accuracy.

%%%%%%%%%%%%%%%%%%%%%%%%%%%%%%%%%%%%%%%%%%%%%%%%%%%%%%%%%%%%%%%%%

\vspace{0.3cm}
\section{Stereo Event Lifetime and Disparity Estimation} \label{sec:method}

%\begin{figure*}[!t]
%	\centering
%	\includegraphics[width=0.8\textwidth]{figures/fit1.png}
%	\caption{}
%	\label{fig:fig2}
%\end{figure*}

In this section we propose a method for estimating the lifetimes of events from two dynamic vision sensors in stereo settings.
Naturally, streams of events emitted by the two sensors are very similar since they mostly observe the same scene.
For the same brightness change, both sensors are triggered and emit an event.
Therefore, it makes sense to calculate the lifetime of those events only once per brightness change and associate it with the other pertaining event as well.

%\begin{figure}[h]
%	\centering
%	\includegraphics[width=0.3\textwidth]{figures/slika00001274.jpg}
%	\caption{}
%	\label{fig:disp}
%\end{figure}

As Algorithm \ref{algo:stereolt} shows, for the current event, we attempt to find its counterpart that was caused by the same brightness gradient in the SAE of the other sensor, which is equivalent to finding its disparity.
We calculate the disparity for each event \emph{asynchronously}, at the time of its arrival.
Even though our approach is event-driven, the method is similar to the standard stereo matching methods in a sense that it operates on frames of accumulated events.
We use an event feature descriptor introduced by \cite{zou2017disparity} that supports event-driven stereo matching based on distance transform \cite{felzenszwalb2012distance} which describes the context of a pixel in binary images by calculating the distance from the pixel to the nearest active pixel.
To apply it to the problem of an event descriptor, coordinates of active events at the time of the current event's arrival are considered as active pixels, and others as inactive.
It is important to note that we make an assumption that the current event is certainly active at its starting point.
Also, even if some future events (relative to the current event) from the other sensor have already been processed, we cannot consider them as active since their starting timestamp is later than the timestamp of the current event.
As a starting step of descriptor computation, we calculate the vector pointing to the nearest active event for every pixel in a $N \times N$ window centered at the current event's position $\mathbf{p}$, separately for events of positive and negative polarity. Then, we classify the obtained vectors by orientation. We assign each vector to one of the 12 bins of the histogram, each bin covering a $30^{\circ}$ interval. The final feature descriptor $\mathbf{D}$ of an event is then defined as a vector of $24$ values; $12$ bins of the histogram for the positive, and likewise for the negative events. Although \cite{zou2017disparity} proposed a descriptor invariant to scale and rotation, we do not implement those features since they are not needed for stereo matching on rectified images.
\begin{algorithm}[!t]
\floatname{algorithm}{Algorithm}
\caption{Stereo event lifetime ($sensor_A$, $sensor_B$)}
\label{algo:stereolt}
\begin{algorithmic}
\STATE{read set of events $E_A$ from $sensor_A$}
\FOR{$e_A(x_i, y_i, t_i)$ in $E_A$}
%	\STATE{$e_A.lifetime = 0$}
	\IF{event $e_B(\_, \_, t_i$) present in $E_B$}
		\STATE{$d$ = computeDisparity($x_i, y_i, t_i$)}
		\IF{$d > 0$}
			\STATE{$lifetime$ = medianLifetime($e_B)$}
		\ELSE
			\STATE{$lifetime$ = computeLifetime($e_A$)}
		\ENDIF
	\ELSE
		\STATE{$lifetime$ = computeLifetime($e_A$)}
	\ENDIF
	\STATE{setLifetime($e_A, lifetime$)}
\ENDFOR
\end{algorithmic}
\end{algorithm}

Disparity map is computed in three standard stereo matching steps: \textit{i)} matching cost calculation, \textit{ii)} cost aggregation, and \textit{iii)} disparity estimation.
Matching cost of an event in location $\mathbf{p}$ for disparity $d$ is calculated by summing absolute differences between the corresponding event descriptors $\mathbf{D}^L$ and $\mathbf{D}^R$, from left and right cameras respectively, within a rectangular window $W_p$ centered at $\mathbf{p}$:
\begin{equation}
C(\mathbf{p}, d)=\sum_{(x_i, y_i) \in W_{p}}\left|\mathbf{D}^{L}(x_i, y_i)-\mathbf{D}^{R}(x_i - d, y_i)\right|.
\end{equation}
Cost aggregation is performed as suggested in \cite{zou2017disparity}, in a fixed region $A$ around the location $\mathbf{p}$:
\begin{equation}
C_{A}^{0}(\mathbf{p}, d)=C(\mathbf{p}, d),\ C_{A}^{i}(\mathbf{p}, d)=\sum_{\mathbf{q} \in A_{p}} C_{A}^{i-1}(\mathbf{q}, d).
\end{equation}
Finally, disparity estimation is done through the winner--takes--all approach, by finding the disparity $d$ with the  minimal aggregated cost:
\begin{equation}
D(\mathbf{p})=\underset{\mathbf{d}}{\arg \min }\ C_{A}(\mathbf{p}, d).
\end{equation}
In case of successful disparity estimation, we determine the current event's lifetime as a median of lifetimes of the events in the matched spatial window $M$:
\begin{equation}
\tau(\mathbf{p}) = \mathrm{median} \{ \tau(\mathbf{q}), \mathbf{q} \in M \}.
\end{equation}
Other computationally light operations for calculating the lifetime using the matched window can be considered as well, e.g., averaging, maximum, minimum, or similar.
We chose median because it is least prone to outliers.
In case of failed disparity estimation, we estimate the lifetime for the event using the aforementioned method.

There are two main advantages of this method opposed to calculating the lifetimes of the events from the two sensors separately.
First, this method is less computationally complex compared to performing the RANSAC algorithm for plane fitting for each event, as will demonstrated in the Section~\ref{sec:exp}.
Furthermore, due to noise and low resolution of the sensor, it is not possible to determine the lifetime of each event using the plane fitting method, so they are falsely declared as noise.
In that case, however, it might be possible to determine the lifetime of the corresponding event in the other sensor's SAE, allowing to retain the event that would otherwise be falsely proclaimed as noise, thus generating a more finely detailed image with sharper gradients.
The method is suitable for execution in a single thread, making it convenient to implement in systems with limited computational resources.
It is also important to note that our method offers high level of abstraction, i.e., it is possible to use any suitable lifetime estimation and disparity estimation method other than the proposed ones depending on the application.

%%%%%%%%%%%%%%%%%%%%%%%%%%%%%%%%%%%%%%%%%%%%%%%%%%%%%%%%%%%%%%%%%%

\vspace{0.3cm}
\section{Experimental Results} \label{sec:exp}

We evaluate the proposed method on multiple real-world stereo experiments.
Our algorithm's output consists of the disparity map and the image of accumulated events for each of the sensor views.
Since there is no metric defined to evaluate the accuracy of lifetime estimation, we compare the outputs qualitatively.
We compare the proposed method in terms of both execution time and accuracy with the method proposed in \cite{zou2017disparity} that first executes lifetime estimation separately for both sensors and then follows up with disparity estimation (referred to as the \textit{decoupled method}). %\im{Referirati se na članak iz kojeg je ta metoda. Daje snagu kada se uspoređuješ s nekim drugim.}
We also compare our output with the method that uses a fixed event accumulation interval, i.e., does not estimate event lifetime.
As future work, the proposed method can be evaluated on the complex real-world stereo dynamic vision sensor dataset \cite{zhu2018multivehicle}.

% \begin{figure}[t]
% 	\centering
% 	\includegraphics[width=0.4\textwidth]{figures/stereo.png}
% 	\caption{Experimental setup. Two DAVIS240C sensors are mounted on a stereo rig with a baseline of $10$ cm. Resolution of the sensors is $240 \times 180$ pixels.}
% 	\label{fig:rig}
% \end{figure}

\subsection{Experimental Setup}

Experiments were conducted with two DAVIS240 sensors mounted on a stereo rig with a baseline of 10 cm as shown in Fig.~\ref{fig:sfig4}.
Resolution of the sensors is $240 \times 180$ pixels.
Timestamps of the events were synchronized on the hardware level.
Event data was undistorted and stereo rectified using the Kalibr toolbox \cite{maye2013self}.
We used the DAVIS240 standard camera frames for calibration, but our proposed method operates on events only.
Experiments were run on a personal computer with Intel Core i7-7700HQ CPU at 2.80GHz on a single core.

\subsection{Implementation Details}
The implementation of the proposed method is done in C++ in ROS environment \cite{quigley2009ros}.
Stream of events from DAVIS240 is fetched in grouped sequences through a ROS topic.
A single thread is subscribed to messages from both left and right sensor.
Random scheduling of the incoming ROS messages does not guarantee that the events from one sensor will always be available before the corresponding events from the other sensor, making it impossible to set one sensor as primary, and other as secondary.
However, for the current set of events, it is possible to determine whether the corresponding events from the other sensor have already been processed and then decide whether to perform lifetime or disparity estimation first.

\begin{figure}[!t]
	\centering
	\vspace{-5mm}
\begin{subfigure}{.24\textwidth}
  \centering
  \includegraphics[width=.99\textwidth]{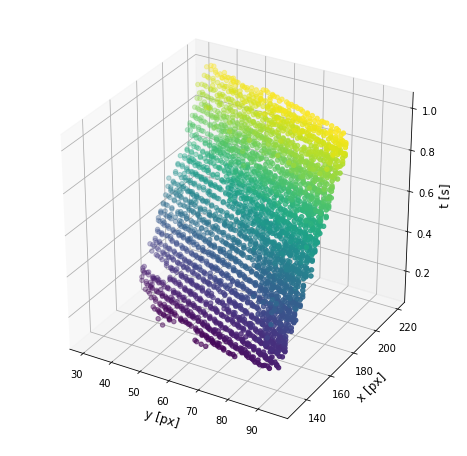}
  \caption{Decoupled, left camera}
  %\label{fig:sfig1}
\end{subfigure}
\begin{subfigure}{.24\textwidth}
  \centering
  \includegraphics[width=.99\textwidth]{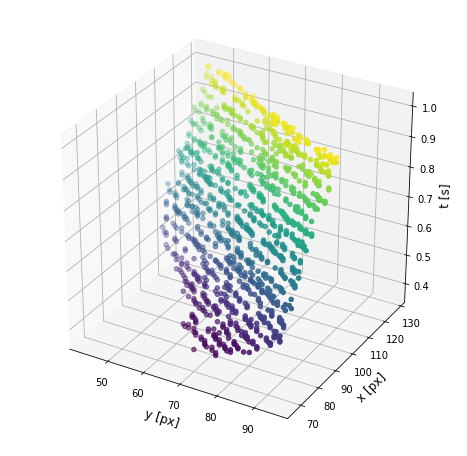}
  \caption{Decoupled, right camera}
  %\label{fig:sfig2}
\end{subfigure}
\begin{subfigure}{.24\textwidth}
  \centering
  \includegraphics[width=.99\textwidth]{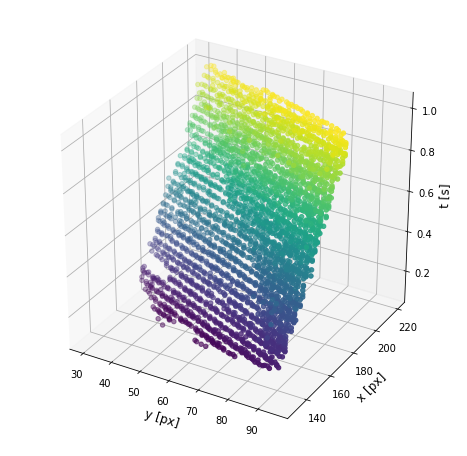}
  \caption{Proposed, left camera}
  %\label{fig:sfig1}
\end{subfigure}%
\begin{subfigure}{.24\textwidth}
  \centering
  \includegraphics[width=.99\textwidth]{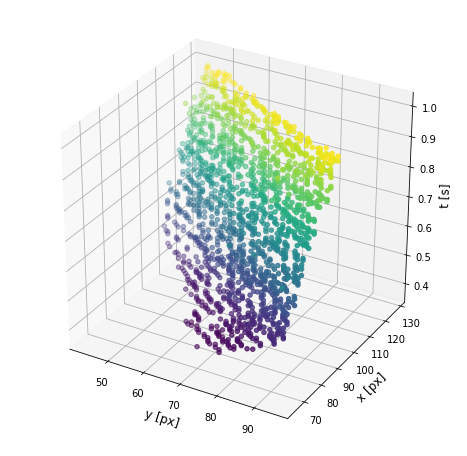}
  \caption{Proposed, right camera}
  %\label{fig:sfig2}
\end{subfigure}

	\caption{SAEs of the events with successfully estimated lifetime obtained by observing a moving line (all events shown in Figure~\ref{fig:sae}). Decoupled method falsely detects many events of the right camera as noise due to sparsity. The proposed method is able to provide lifetime for more events (1642) than the decoupled method (1178), thus providing better accuracy for the right camera, while retaining accuracy for the left camera.}
	\label{fig:accuracy}
	\vspace{-2mm}
\end{figure}

\subsection{Accuracy Improvement}

\begin{figure}[!t]
\begin{subfigure}{.24\textwidth}
  \centering
  \frame{\includegraphics[width=.95\textwidth]{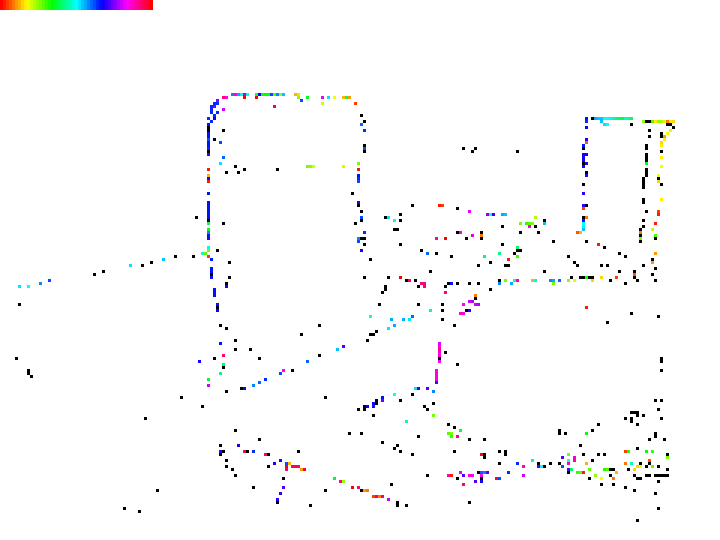}}
  \caption{Fixed accumulation interval}
  %\label{fig:sfig1}
\end{subfigure}%
\begin{subfigure}{.24\textwidth}
  \centering
  \frame{\includegraphics[width=.95\textwidth]{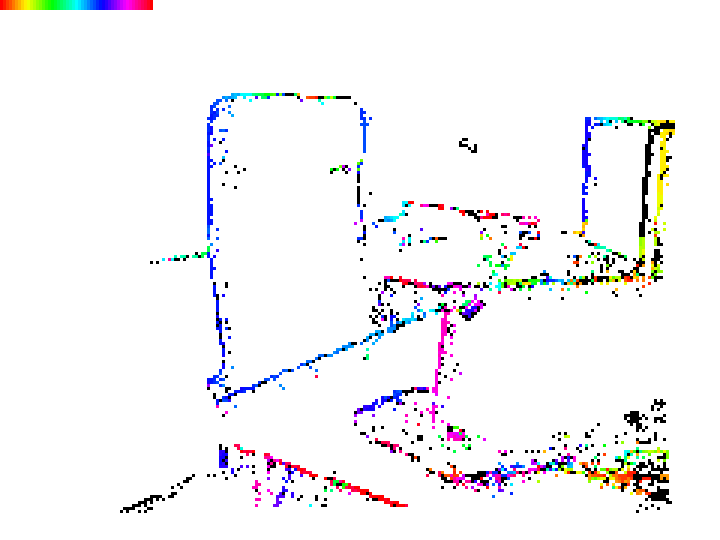}}
  \caption{Proposed method}
 %\label{fig:sfig2}
\end{subfigure}
\caption{Comparison with respect to a fixed accumulation time interval of 10\,ms. Disparity estimation is not reliable for nearly horizontal lines and areas prone to noise due to lack of structure, but our method still yields sharp gradient images.}
\label{fig:exp3}
\end{figure}

\begin{figure}[!t]
\begin{subfigure}{.24\textwidth}
  \centering
  \frame{\includegraphics[width=.95\textwidth]{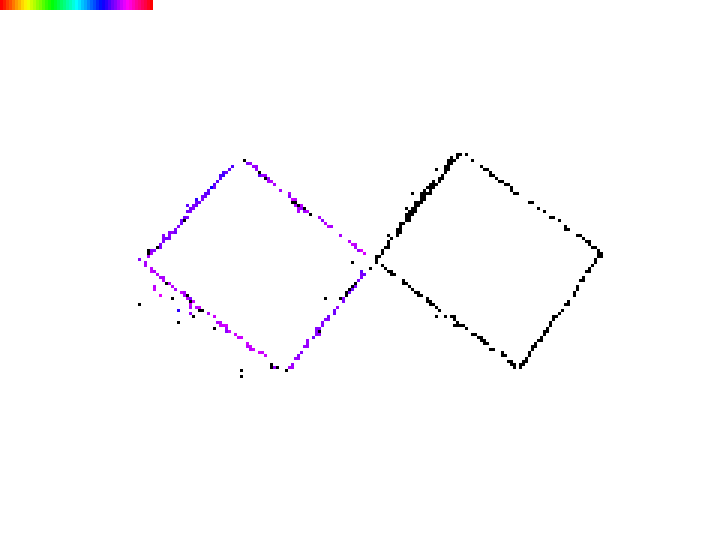}}
  \caption{Decoupled method}
  %\label{fig:sfig1}
\end{subfigure}%
\begin{subfigure}{.24\textwidth}
  \centering
  \frame{\includegraphics[width=.95\textwidth]{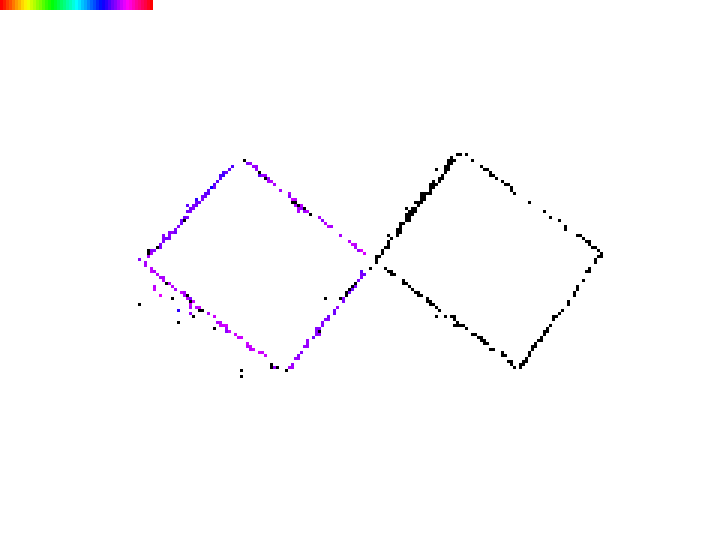}}
  \caption{Proposed method}
  %\label{fig:sfig2}
\end{subfigure}
\caption{Image shows events from both sensors. Although both decoupled and proposed method yield sharp gradients and correct disparity in this example, our method performs nearly twice as fast compared to the decoupled approach.}
\label{fig:exp1}
%\vspace{-5mm}
\end{figure}

DVS yield sparse event data in case of a motion occurring within the area of very similar brightness intensity, which the sensor is unable to discern, resulting in poor results of event lifetime and disparity estimation.
Figure~\ref{fig:accuracy} shows events obtained by recording a moving line from a computer screen.
Due to sparse event data caused by screen flickering, the decoupled lifetime estimation method was not able to compute the lifetime of some events from the right sensor, thus proclaiming many events as noise.
However, as the two sensors are naturally not completely identical in terms of sensitivity to brightness changes, the method was more successful in computing the lifetimes of events from the left sensor.
Since the proposed stereo event lifetime computation method also uses events from the left sensor, it was able to compute lifetimes for more events from the right sensor compared to the decoupled lifetime computation, thus proving better accuracy in terms of noise detection compared to the decoupled method.
%\im{Opisati rezultate iz slika 5 i 6}\im{discuss}
Experiment shown in Figure~\ref{fig:exp1} confirms that our method also matches the decoupled method in terms of accuracy in case of dense event data.
Figures~\ref{fig:exp2} and \ref{fig:exp3} show the experiments performed on real-world data comparing the proposed method to using a fixed event accumulation time interval of 10\,ms, which yields very sparse gradient image lacking structure. 

\subsection{Execution Time Improvement}

Since the plane-fitting algorithm used to estimate the event lifetime is computationally complex, our method aims to reduce the number of times it needs to be executed as much as possible.
As a first way of reducing execution time, we propose to use the previously calculated and predicted lifetime, if the prediction was reliable, without fitting a new plane.
Furthermore, if the predicted lifetime is not available, our method executes plane-fitting only once for the two corresponding events caused by one brightness change, effectively cutting the execution time in half.
The amount of improvement depends on the number of events whose disparity is successfully estimated, in which case we use the efficient method of calculating the current event lifetime by finding the median of lifetimes of the events in the matched window.
Experiments show that the execution time is reduced approximately by 50\% (see Table~\ref{table:results}).

\begin{table}[!t]
\caption{Number of events with estimated lifetime and pertaining estimation execution times. The proposed method runs nearly twice as fast compared to the decoupled method. The time needed to calculate the disparity is not included.}
\begin{center}
\begin{tabular}{  c  c c c c }
 \hline
 & \multicolumn{2}{c}{Decoupled} & \multicolumn{2}{c}{Proposed} \\
 \cline{2-5}
 \multicolumn{1}{c}{Scene} & Events & Time [ms] & Events & Time [ms] \\ \hline
 Figure \ref{fig:exp2} & 2692 & 32 & 2759 & 16 \\
 Figure \ref{fig:exp3} & 2536 & 30 & 2498 & 15 \\
 Figure \ref{fig:exp1} & 426 & 11 & 436 & 6 \\
 \hline
\end{tabular}
\end{center}
\label{table:results}
\end{table}

\vspace{0.3cm}
\section{Conclusion} \label{sec:conclusion}

In this paper, we have presented a novel method for stereo lifetime and disparity estimation for event-based cameras.
Enhancing the events with their lifetime prior to disparity estimation makes the disparity maps representative of the scene structure invariant to the amount of motion of the sensor or the scene.
Since one brightness change triggers an event in both of the sensors, we have proposed to calculate the lifetime for only one of those events, and associate it to the other one via stereo matching, while simultaneously determining its disparity.
The proposed method is twice as fast as the method calculating the event lifetimes for each sensor independently, while it also provides disparity maps with more accurate structure, as we have demonstrated through several real-world experiments.
% Future work will include improving lifetime estimation method by introducing the inertial measurement unit and considering other types of descriptors for stereo matching.

\balance

%\vspace{2cm}
\bibliography{main}
\bibliographystyle{IEEEtran}

\end{document}